%% file: main.tex
\documentclass[sigconf]{acmart}

\usepackage{comment}

\usepackage{epigraph}
\makeatletter
\newenvironment{chapquote}[2][2em]
  {\setlength{\@tempdima}{#1}%
   \def\chapquote@author{#2}%
   \parshape 1 \@tempdima \dimexpr\textwidth-2\@tempdima\relax%
   \itshape}
  {\par\normalfont\hfill--\ \chapquote@author\hspace*{\@tempdima}\par\bigskip}
\makeatother

\usepackage{graphicx}

\usepackage{multicol}

\usepackage{ifthen}
\newboolean{techreport}
\usepackage{caption}
\usepackage{multirow}
\usepackage{subcaption}
\usepackage{hyperref}
\usepackage{amsmath}
\usepackage{tikz}
\usepackage[]{footmisc}
\usepackage{graphicx}
\usepackage{pifont}

\usepackage[ruled,linesnumbered,vlined]{algorithm2e}
\let\oldnl\nl
\newcommand{\nonl}{\renewcommand{\nl}{\let\nl\oldnl}}
\usepackage{tikz}

\definecolor{codegreen}{rgb}{0,0.6,0}
\definecolor{codegray}{rgb}{0.5,0.5,0.5}
\definecolor{codepurple}{HTML}{C42043}
\definecolor{backcolour}{rgb}{1,1,1}

\newcommand\numberstyle[1]{%
    \footnotesize
    \color{codegray}%
    \ttfamily
    \ifnum#1<10 0\fi#1 |%
}

\usepackage[ruled,linesnumbered]{algorithm2e}
\usepackage{algpseudocode}
\SetKw{KwBy}{by}
\SetKwComment{Comment}{/* }{ */}
\RestyleAlgo{ruled}

\usepackage{enumitem}

\renewcommand\footnotetextcopyrightpermission[1]{} 

\begin{document}
\setboolean{techreport}{true}

\title{Safety in the Emerging Holodeck Applications}
\titlenote{Appeared in~\cite{shahram2022b}, the {\em CHI Conference on Human Factors in Computing
Systems (CHI ’22)}, April 29-May 5, 2022, New Orleans, LA, USA. ACM, New York, NY, 3 pages.}

\author{Shahram Ghandeharizadeh, Luis Garcia}
\affiliation{%
  \institution{University of Southern California}
    \city{Los Angeles}
    \state{California}
  \country{USA}
}
\email{shahram@usc.edu, lgarcia@isi.edu}

\begin{abstract}
\input abs
\end{abstract}

\maketitle
\pagestyle{plain} 

\input introduction

\input future

\bibliographystyle{ACM-Reference-Format}
\bibliography{main}  

\end{document}

%% file: abs.tex
Technological advances in holography, robotics, and 3D printing are starting to realize the vision of a holodeck.
These immersive 3D displays must address user safety from the start to be viable.
A holodeck’s safety challenges are novel because its applications will involve explicit physical interactions between humans and synthesized 3D objects and experiences in real-time. 
This pioneering paper first proposes research directions for modeling safety in future holodeck applications from traditional physical human-robot interaction modeling. 
Subsequently, 
we propose a 
test-bed to enable safety validation of physical human-robot interaction based on existing augmented reality and virtual simulation technology.

%% file: introduction.tex
\section{Introduction}


\begin{chapquote}
{\small``The ultimate display would, of course, be a room within which the computer can control the existence of matter.''~\cite{sutherland1965}}
\end{chapquote}

A holodeck enables users to see virtual objects without glasses and to interact with them without wearing gloves or bodysuits.  Holodecks may occupy physical volumes such as a tabletop cuboid or sphere, a telephone booth, a self-driving vehicle, a room, a concert hall, or a stadium.  Holodecks may be realized via holograms~\cite{hopper2000reality}, 2D screens~\cite{vipe2014}, a ground simulator~\cite{mario2020Walking}, fast 3D printing~\cite{t1000}, claytronics as physical artifacts using programmable matter consisting of catoms~\cite{matter2005}, roboxels as cellular robots that dynamically configure themselves into the desired shape and size~\cite{roboxel1993}, BitDrones as interactive nano-quadcopter~\cite{gomes2016bitdrones}, Flying Light Specks (FLSs) as miniature drones with Red/Green/Blue (RGB) lights that fly as swarms to illuminate a virtual object~\cite{ghandeharizadeh2021holodeck,shahram2022}, or any combination of these.  
With 3D printing, claytronics, roboxels, BitDrones and FLSs, a holodeck facilitates encounter-type haptic displays~\cite{rodrigo2021} by materializing a synthesized physical representation of a virtual object, enabling one or more users to interact with (e.g., touch) them in the physical world.  These physical objects may have behavior that enables them to interact with users---potentially in a precarious manner.  

\noindent\textbf{Example use case: fencing in the holodeck.} Consider the fencing mixed reality application proposed by the industry~\cite{kraus2022facebook,won2021control} realized in the holodeck, e.g., a training session in a holodeck between a user and their trainer.  Each may be in a different specialized room, a holodeck, in different cities.  The 500-gram saber used by each user may be 3D printed and configured with FLSs as its sensors.  With the user’s (trainer’s) holodeck, the trainer (user) and their sword may be illuminated as FLSs.  The FLSs will repel one another using heartbeat messages~\cite{ghandeharizadeh2021holodeck}, which are used to detect when the 3D printed saber scores a touch without damaging the FLSs.  For example, in the user’s holodeck, when the user lands a scoring touch on the FLS illuminated trainer, the FLSs that constitute the trainer go dark and disperse by detecting the in-coming heartbeat messages of the FLSs in the user’s saber. This process occurs prior to the saber colliding with the FLSs, preventing damage to FLSs.


The inherent safety challenge resides in protecting the user from injury while maintaining the utility and realism of the fencing application. A possible approach is to require the application to be intrinsically safe, e.g., requiring participants
to wear specialized protective clothing 
with sensors. Alternatively, the holodeck can employ fine-grained tracking to either avoid collisions, e.g., detecting when an FLS illuminated saber is about to touch a user and have its FLSs 
disperse, or control the severity of impact, e.g., allow the user to experience the impact of a saber by having the FLSs exert sufficient force. 

\noindent\textbf{Research challenges.} Inspired by this fencing scenario, we summarize a non-comprehensive set of research challenges for holodeck safety as follows.
\begin{itemize}
    \item \textit{Challenge 1:} How can a holodeck ensure its safety functionality is implemented correctly and prove that its execution does not violate its safeguards?
    \item \textit{Challenge 2:} How can a synthesized physical object exert force on a user's skin without causing injury to the user while providing the appropriate haptic feedback?
    \item \textit{Challenge 3:} What are the short-term and long-term safety hazards associated with the use of FLSs and chemical processes used by fast 3D printers?
    \item \textit{Challenge 4:} How can physical assets that may cause physical harm be constrained to the holodeck environment?
    \item \textit{Challenge 5:} What are the safety issues with a user falling down on a holodeck floor equipped with a ground simulator~\cite{mario2020Walking}?  When and how should the fall be prevented?
\end{itemize}

Since the technology has not been fully realized, there is currently no science to address these safety challenges explicitly. The science is critical to ensure the viability of emerging holodeck applications. Otherwise, an accidental injury may be a missed scientific opportunity to enhance safety of holodecks.  At their worst, accidents may motivate government involvement and legislation that hinder or completely halt the research and development of the holodeck technology. 

In addition, both the security of holodecks and the privacy of their users are significant topics that relate to safety.  We defer both to future work in order to maintain focus on safety with no adversary. 

The {\em contributions} of this paper consist of:
\begin{enumerate}
    \item identifying the role of tolerance metrics as the key to enhancing human physical interaction with synthesized 3D objects in a holodeck, and 
    \item design goals of a testbed for safe implementation of touch sensation using FLSs.
\end{enumerate}

%% file: future.tex
\section{Modeling Physical Human-Robot Interaction Safety in the holodeck}\label{sec:humanrobot}
Traditional physical human-robot interaction safety focused on developing intrinsically safe applications by segregating human and robot interaction altogether, e.g., by avoiding interactions or collisions between a human and the dangerous robot components~\cite{colgate2008safety}. Of course, with the advent of intelligent assistant robots, human-robot interaction became more physically explicit and collaborative.
There is a trade-off between safety and application realism.
For instance, a user may be willing to endure some physical pain during a fencing session 
with a synthesized physical avatar.
A future holodeck application must quantify the tolerable level of discomfort and communicate it to its user intuitively.
A solution may calibrate a user and tailor the holodeck application for each individual.

Several domain-specific metrics and standards have been proposed to quantify user safety when explicitly collaborating with a robot in traditional physical human-robot interaction safety literature. Generally, researchers have proposed tolerance measures for physical forces applied to specific human body areas. For instance, the head injury criterion (HIC) is a severity index that bounds the duration and acceleration force applied to a human head in automotive applications~\cite{versace1971review}. 
Similar tolerance metrics must be developed for different modes of operation in holodeck applications. 
Moreover, application developers must  communicate the physical risks associated with their application through standardized means. 
The enabling holodeck technology should also vet third-party applications against their specified risks and interface these risks efficiently with the user. 
Finally, human-in-the-loop robot interaction introduces a significant amount of uncertainty into applications modeling physical interaction~\cite{axelrod2018provably}. 

\section{Enabling Safety Validation for Future Holodeck Applications}\label{sec:safety}
While holodeck applications need to be safe, it is equally important that they do not operate so conservatively to sacrifice their quality of realism (QoR). Thus, a scientific framework is necessary to detect and identify safety risks, root causes, and how an application may resolve them while maximizing its QoR. Concretely, a testbed is needed to standardize formal methods and metrics for communicating to the stakeholders when an application incurs safety risks, identifying the risk and how it compromises human safety, and how the application may address the risk while maximizing its QoR. The testbed should track all physical and virtual objects at a fine granularity, using both first-order principles and data-driven approaches for its analysis. The testbed will include sufficient computation for simulation, physical state estimation, and runtime verification of a synthesized physical object (e.g., a saber) or experience (e.g., walking back and forth). In addition, it will support diverse applications by generalizing the safety risk of one across others to evaluate its feasibility in other applications. 
This enables the testbed to prevent other applications from encountering the same risk.
Such a testbed enhances user safety through standards, education, and research.
Below, we provide an overview of our immediate short-term plans towards a testbed. 

\noindent\textbf{Simulating the application developer's perspective in the holodeck.} Validating the safety of an application relative to a user model can be either simulated physically or digitally, e.g., having an actual human interacting with digital objects or simulating a data-driven human model. For human physical modeling, the users would be equipped with fine-grained motion capture technology to perform pose estimation and evaluate various collision contingencies with digital objects. Conversely, human motion models from existing datasets or simulators can be utilized to generate a copious amount of possible interactions with digital objects. Moreover, the human motion models could interact with synthesized physical objects stemming from the holodeck technology to evaluate novel scenarios' safety. Finally, such a testbed would enable validation of the real-time constraints across the computation required for perception, cognition, and control.

\noindent\textbf{Simulating a user's perspective in the holodeck.} Safely validating a user's safety against a holodeck application would be no different from a simulated augmented reality experience. The user would ideally not wear any wearable technology in the holodeck environment. However, to ground the testbed in existing technology, the user would be provided an augmented reality experience with haptic feedback to simulate physical objects. 

\noindent\textbf{Progressing from virtual to physical simulated holodeck objects.} 
We propose to progressively integrate synthesized, physical objects that are intrinsically safe concerning human models. 
In today's technology, synthesizing micrometer FLSs is not feasible because none have been developed as yet. However, scaled-up experiments using today's small drones (tens of milimeter in size costing \$20) can be executed to simulate a subset of the interactions analogously
relative to a human model.
A functionality is human touch using FLS-matter, 2D and 3D geometric shapes using FLSs~\cite{ghandeharizadeh2021holodeck}.  These shapes will have several programmed behaviors that react to human touch.  These include a shape moving along a trajectory as a function of the force exerted by a user to its side, move up when a user exerts force to the bottom of it exceeding gravity pull, move back to its original position when the user stops exerting force, and disintegrate when the user exerted force exceeds a certain threshold.  With each, the swarm of FLSs will exert some force back at the point of human finger perturbation to provide the touch sensation. We conduct user studies with these shapes. 
Our objective is to quantify safety risks and how realistic the touch sensation of the FLS-matter feels to the users.  We will use surveys to document user experiences and QoR.

The testbed and its obtained results will provide insights into safety risks and QoR trade-off as new holodeck technology (e.g., smaller-sized FLSs) become available.